%% file: root.tex
\documentclass[letterpaper, 10 pt, conference]{ieeeconf}  % Comment this line out if you need a4paper

\IEEEoverridecommandlockouts                              % This command is only needed if 
\usepackage{graphics} % for pdf, bitmapped graphics files
\usepackage{epsfig} % for postscript graphics files
\usepackage{times} % assumes new font selection scheme installed
\usepackage{amsmath} % assumes amsmath package installed
\usepackage{amssymb}  % assumes amsmath package installed
\usepackage{caption}
\usepackage{subcaption}
\usepackage{cite}
\usepackage{hyperref}

\title{\LARGE \bf
An Attention-based Recurrent Convolutional Network \\
for Vehicle Taillight Recognition
}

\author{Kuan-Hui Lee, Takaaki Tagawa, Jia-En M. Pan, Adrien Gaidon, Bertrand Douillard% <-this % stops a space
\thanks{All the authors are with Toyota Research Institute, CA USA. {\tt\small \string{kuan.lee, takaaki.tagawa, marcus.pan, adrien.gaidon, bertrand.douillard\string}@tri.global}}%
}

\begin{document}

\maketitle
\thispagestyle{empty}
\pagestyle{empty}

%%%%%%%%%%%%%%%%%%%%%%%%%%%%%%%%%%%%%%%%%%%%%%%%%%%%%%%%%%%%%%%%%%%%%%%%%%%%%%%%

\input{0.abstract.tex}

\input{1.introduction.tex}

\input{2.related_work.tex}

\input{3.method.tex}

\input{4.experiments.tex}

\input{5.conclusions.tex}

% \addtolength{\textheight}{-12cm}  % This command serves to balance the column lengths

%%%%%%%%%%%%%%%%%%%%%%%%%%%%%%%%%%%%%%%%%%%%%%%%%%%%%%%%%%%%%%%%%%%%%%%%%%%%%%%%

\bibliographystyle{IEEEtran}
\bibliography{IEEEabrv,reference}

\end{document}

%% file: 0.abstract.tex
\begin{abstract}

Vehicle taillight recognition is an important application for automated driving, especially for intent prediction of ado vehicles and trajectory planning of the ego vehicle.
In this work, we propose an end-to-end deep learning framework to recognize taillights, i.e. rear turn and brake signals, from a sequence of images.
The proposed method starts with a Convolutional Neural Network (CNN) to extract spatial features, and then applies a Long Short-Term Memory network (LSTM) to learn temporal dependencies.
Furthermore, we integrate attention models in both spatial and temporal domains, where the attention models learn to selectively focus on both spatial and temporal features.
Our method is able to outperform the state of the art in terms of accuracy on the UC Merced Vehicle Rear Signal Dataset, demonstrating the effectiveness of attention models for vehicle taillight recognition.

\end{abstract}

%% file: 1.introduction.tex
\section{INTRODUCTION}
\label{sec:introduction}

% Autonomous driving is one of the most popular applications in academic, commercial, and public communities.
%
Since the 2005 DARPA Grand Challenge and the 2007 Urban Challenge, researchers have developed many successful technologies towards automated driving, including key perception tasks \cite{leonard2008perception,lee2018multi,badue2019self}.   
Nonetheless, higher level concepts such as intent prediction, human-machine interaction, and vehicle-to-vehicle communication, are still open questions.
In particular, intent prediction of ado vehicles is one of the most critical features for automated driving safety.

In this work, we propose a method to perceive one of the key explicit intent signals the ego vehicle needs to understand from the ado vehicles: the status of turn and brake lights.
Numerous methods have been proposed to recognize vehicle taillight signals.
Typical computer vision pipelines \cite{wang2016appearance,hsu2017learning,thammakaroon2009predictive,chen2016daytime,chen2014nighttime,almagambetov2012autonomous,casares2012robust,cui2015vision,zhong2016learning} mainly detect and extract ado vehicles' bounding boxes in the frames captured by the ego vehicle's front-facing camera, and classify recognize the signal states.
This also requires encoding temporal dependencies since the signal states are determined by changes of activation of the lights.
Hence, vehicle taillight recognition can be treated as an application of video recognition.
This enables leveraging progress in related deep learning models, especially in action recognition or image caption generation.
To tackle the problem, there are two main types of methods based on the network model: the Recurrent Neural Network (RNN) based methods and the 3-D CNN based methods.  
The RNN based methods \cite{donahue2015long,zhu2016co,gammulle2017two} usually use the CNN for feature extraction and apply the features to a recurrent network, such as the LSTM.
The 3-D CNN based methods \cite{tran2015learning}\cite{carreira2017quo} apply convolution operations not only to the spatial domain but also to the temporal domain.
Besides, facilitated by optical flow, a two-stream fusion architecture is proposed to fuse spatial and temporal dependencies together \cite{simonyan2014two}\cite{feichtenhofer2016convolutional}.
Such deep learning methods have achieved promising results, inspiring the researchers to apply the ideas and the techniques to vehicle taillight recognition \cite{wang2016appearance}\cite{hsu2017learning}.

% For instance, Donahue et al. \cite{donahue2015long} proposed Long-term Recurrent Convolutional Networks (LRCN), which combines CNN and LSTM network to generalize network architecture for visual recognition and description.
%
%In \cite{feichtenhofer2016convolutional}, a two-stream fusion architecture was proposed to leverage optical flow as another stream and eventually fuse spatial and temporal dependencies together.
%
% Other than the LSTM based method, Tran et al. \cite{tran2015learning} proposed a Convolutional 3-D (C3D) network to encode temporal dependencies by means of 3-D convolution.  
%
% Carreira et al. \cite{carreira2017quo} proposed two-stream Inflated 3-D ConvNets (I3D), which combines 3-D based convolution into two stream architecture with inflated filters and kernels.
%

In this paper, we leverage similar insights and propose a novel end-to-end recurrent convolutional neural network for vehicle taillight recognition.
Our model is based on the CNN-LSTM framework paired with a spatio-temporal attention model that emphasizes not only focal regions of the images but also key time steps of the sequences.
As a result, the proposed method is able to outperform the state of the art, achieving better accuracy performance.
Furthermore, our model is more interpretable, as our spatio-temporal attention maps allow us to verify whether the inference relied on appropriate causal factors (spatio-temporal regions) for its prediction.
In summary, our main contributions are:
\begin{itemize}
\item We propose an end-to-end deep neural network for vehicle taillight recognition, where the architecture is mainly based on the CNN-LSTM framework. 
\item The proposed method integrates a spatial attention model into the framework, so as to emphasize "regions of interest" of the images.
\item Our model also contains a temporal attention model to concentrate on certain time steps that are key to the recognition task.
\end{itemize}

The rest of the paper is organized as follows.
Section \ref{sec:related_work} gives a brief overview of related works.
Section \ref{sec:method} depicts the proposed method and network architectures.
The experimental results are shown in Section \ref{sec:experiments}, followed by the conclusion in Section \ref{sec:conclusions}.

%% file: 2.related_work.tex
\section{RELATED WORK}
\label{sec:related_work}

\begin{figure*}[ht!]
\centering
\includegraphics[width=\textwidth]{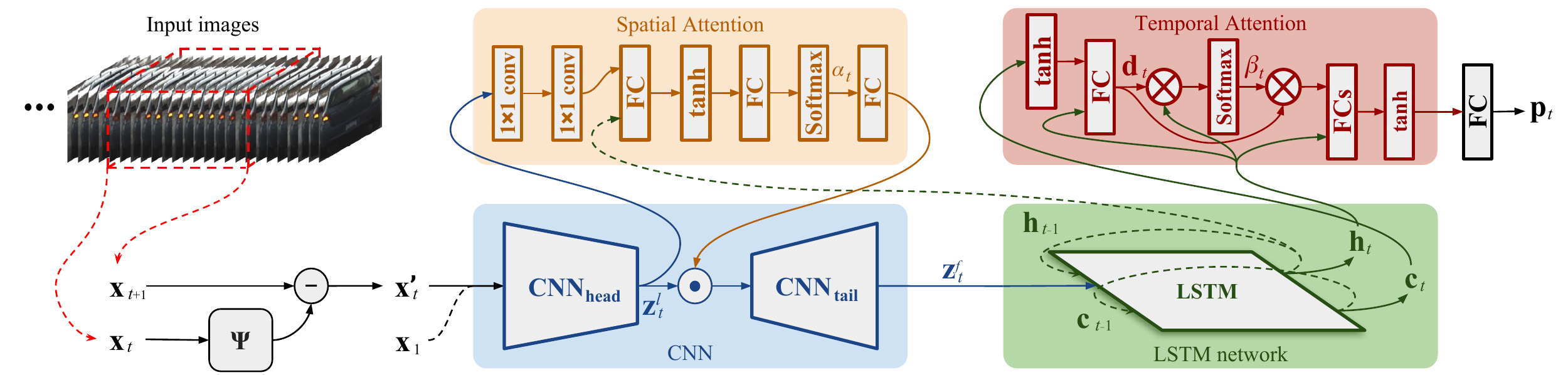}
\caption{
Network architecture. The input images are a sequence of one vehicle's bounding boxes. The absolute difference is obtained from two successive images. The spatial attention is applied to the  $l^\text{th}$ layer's features, and the temporal attention is applied to the hidden variables of the LSTM network. Here $\odot$ is the element-wise product and $\otimes$ is the matrix product.}
\label{fig:overview}
\vspace{-4mm}
\end{figure*}

\subsection{Vehicle Taillight Recognition}

Several taillight recognition methods detect the signal states by thresholding on color features \cite{thammakaroon2009predictive}\cite{ chen2016daytime}.
Chen et al. \cite{chen2014nighttime} use color thresholds to train an AdaBoost based classifier for the existence of the turn signals, and then use reflectance contrast to tell the directions.
Frohlich et al. \cite{frohlich2014will} localize the turn light regions according to absolute difference between two successive frames, and train an AdaBoost based classifier for the extracted features which is transferred to the frequency domain.
Chen et al. \cite{chen2015robust} propose a response function to determine the taillights regions.
A high-pass mask is used to find the brighter in-region pixels for classifying the states accordingly.
Several methods \cite{almagambetov2012autonomous, casares2012robust, almagambetov2015robust} utilize tracking to localize taillights.
The states are classified based on the luminance channel observed over time on the two detected light regions.
Cui et al. \cite{cui2015vision} detect the turn lights based on thresholds, and use support vector machine (SVM) to classify signals into four states.

Recently, deep learning approaches have also been applied to learn features for taillights.
Zhong et al. \cite{zhong2016learning} train a fully convolutional network (FCN) \cite{long2015fully} model to identify the light regions and the features extracted within the regions are classified by a linear SVM.
Wang et al. \cite{wang2016appearance} train a CNN model from vehicle rear appearances to tell the state of the brake signals image by image.
In order to take temporal dependencies into account, Hsu et al. \cite{hsu2017learning} propose a CNN-LSTM structure to learn eight states of taillights, where the networks for brake and turn signals are trained separately.

%%%%%%%%%%%%%%%%%%%%%%%%%%%%%%%%%%%%%%%%%%%%%%%%%%%%%%%%%%%%%%%%%%%%%%%%
\subsection{Attention-Based Models}

Neural attention models focus on certain key parts of the data to improve inference.
Soft-selection has recently proven effective in many recurrent applications, such as machine translation \cite{bahdanau2014neural,luong2015effective,gehring2017convolutional}, image caption generation \cite{xu2015show}, image recognition \cite{ba2014multiple}\cite{mnih2014recurrent}, and human action recognition \cite{sharma2015action, song2017end,yeung2018every}.

In machine translation, attention models are widely applied in the sequence-to-sequence framework proposed in \cite{sutskever2014sequence}. Bahdanau et al. \cite{bahdanau2014neural} apply a soft-alignment mechanism to jointly translate and selectively align words.
Luong et al \cite{luong2015effective} propose local and global attention mechanisms where the global (resp. local) approach attends to the whole (resp. a subset of the) sequence.
Gehring et al. \cite{gehring2017convolutional} introduce an architecture with CNNs to optimize GPU usage. The proposed method also equips each decoder layer with a separate attention model.

Attention mechanisms have also proven effective for a wide array of computer vision tasks.
In \cite{xu2015show}, Xu et al. adapt the attention mechanism to the spatial domain such that the proposed network automatically learns to concentrate on salient objects.
Ba et al. \cite{ba2014multiple} propose a RNN model trained with reinforcement learning to recognize multiple objects in images, where the network focuses on the most relevant regions of the input image.
Mnih et al. \cite{mnih2014recurrent} train a neural network to learn alignments between image objects and agent actions in dynamic control problems.

Beyond text and images, several works also investigate attention models for human action recognition, achieving significant performance improvements~\cite{xu2015show, ba2014multiple, mnih2014recurrent, sharma2015action, yeung2018every, song2017end}.
Yao et al. \cite{yao2015describing} propose a 3-D CNN-RNN encoder-decoder architecture that captures not only local spatio-temporal information but also global context via an attention mechanism. 
Sharma et al. in\cite{sharma2015action} apply a spatial attention model to video frames and embed the attention weights into a multi-layer LSTM network for action recognition.
In \cite{song2017end}, Song et al. propose an end-to-end training framework based on the LSTM model with spatial and temporal attention models for human joints.
Yeung et al. \cite{yeung2018every} take adjacent frames within a sliding window into account and propose a recurrent LSTM-based model, where a temporal attention model is learned to enhance the performance of dense labeling of actions.

%% file: 3.method.tex
\section{PROPOSED FRAMEWORK}
\label{sec:method}

We propose a CNN-LSTM based network with spatial and temporal attention mechanisms for vehicle taillight recognition.
An overview of the architecture is shown in the Figure \ref{fig:overview}.
By using object detection techniques, a sequence of one vehicle's bounding boxes images can be extracted from the video frames.
From the sequence, chunks of the images are sampled by window-sliding along the temporal direction.
Assume $\textbf{X}=\{\textbf{x}_t\}_{t=1}^{T+1}$ is a chunk with $T+1$ images, where $\textbf{x}_t$ is the $t^{\text{th}}$ image in the chunk.
Inspired by \cite{simonyan2014two}\cite{feichtenhofer2016convolutional} which leverage optical flow to present the temporal dependencies, we first calculate the frame difference $\textbf{x}'_t$ by using the SIFT flow algorithm \cite{liu2011sift} to align the vehicles in successive frames:
\begin{equation}
\textbf{x}'_t = |\Psi^{t-1 \rightarrow t}(\textbf{X}_{t-1}) - \textbf{X}_t|,\text{ for } t=1,2,...,T.
\label{eq:sift_flow}
\end{equation}
where $\Psi^{t \rightarrow t+1}(\bullet)$ is a warping function based on their SIFT flow, from the $t^\text{th}$ frame to the next frame.
Combining with $\textbf{x}_{1}$, the input of the network becomes $\textbf{X}'=\{\textbf{x}_1, \textbf{x}'_1,...\textbf{x}'_T\}$.

Each image is forwarded to certain CNN layers ($\text{CNN}_\text{head}$), to obtain deep features in the $l^{\text{th}}$ layers, denoted by $\textbf{Z}^l=\{\textbf{z}^l_t\}_{t=1}^T$.
The input of the spatial attention model, $\textbf{z}^l_t$, is forwarded to two 2-D convolutional layers, and the output is concatenated with the last hidden variable $\textbf{h}_{t-1}$ from the LSTM.
Then, the concatenated tensors are forwarded to the fully connected layers (FC), the hyperbolic tangent (tanh) layer, and the softmax layer, so as to obtain the attention weights $\alpha_t$.
The deep features $\textbf{z}^l_t$ perform element-wise multiplication with $\alpha_t$, and are then forwarded to the rest part of CNN ($\text{CNN}_\text{tail}$) for latent features $\textbf{Z}^f=\{\textbf{z}^f_t\}_{t=1}^T$.
The latent features are then forwarded to a LSTM network for encoding temporal dependencies.

The input and the output of the LSTM network are recurrent over time steps.
At time step $t$, the input is the latent feature $\textbf{z}^f_t$ from the last fully connected layer of the CNN, the hidden variable $\textbf{h}_{t-1}$, and the internal states $\textbf{c}_{t-1}$, while the output is the hidden variable $\textbf{h}_{t}$ and the memory cell $\textbf{c}_{t}$ for the next time step.
Both $\textbf{h}_{t}$ and $\textbf{c}_{t}$ are updated and then passed to the LSTM network at every time step.
%

% The LSTM unit is shown in Figure \ref{fig:lstm}, where the inputs and the outputs are recurrent over time steps.
% %
% At time step $t$, the inputs are the latent feature $\textbf{z}^f_t$ from the last fully connected layer of the CNN, the hidden variable $\textbf{h}_{t-1}$, and the internal states $\textbf{c}_{t-1}$, while the outputs are the hidden variable $\textbf{h}_{t}$ and memory cell $\textbf{c}_{t}$ for the next time step.
% %
% The memory cell is updated at every time step and then passed to the next LSTM unit.
% %
% All the updates and outputs are processed by the different gates which are sigmoid functions that outputs values from 0 to 1.
% %
% First, the forget gate $\textbf{f}_t$ determines what to discard from $\textbf{z}^f_{t}$ and $\textbf{h}_{t-1}$, and what to remember from $\textbf{c}_{t-1}$, by performing element-wise multiplication.
% %
% Next, the LSTM unit updates information to the memory cell through the input gate $\textbf{i}_t$ and the tanh layer $\textbf{g}_t$. These two gates control what information to remember from $\textbf{z}^f_{t}$ and $\textbf{h}_{t-1}$, are then added to the memory cell. , i.e., deep
% %
% Finally, the LSTM unit outputs a hidden state $\textbf{h}_t$ at every time step. The output gate $\textbf{o}_t$ is computed and weighted with the cell states through a tanh layer to determine the hidden variable $\textbf{h}_t$.

The LSTM network outputs a set of hidden variables $\textbf{H}=\{\textbf{h}_t\}_{t=1}^T$ and a set of memory cells $\textbf{C}=\{\textbf{c}_t\}_{t=1}^T$, which are used in temporal attention model. 
The temporal attention model adopts the attention model proposed in \cite{gehring2017convolutional}, which calculates the attention by taking the dot-product between decoder context and encoder representations.
Instead of multiple decoder layers, we only use single layer for the output of the LSTM network.
The input $\textbf{h}_{t}$ and $\textbf{c}_{t}$ are fused to be a set of state summaries $\textbf{D} = \{\textbf{d}_{t}\}^T_{t=1}$.
Then, $\textbf{H}$ and $\textbf{D}$ perform a matrix product, where the results are followed by a softmax layer for attention selection.
The attention weights are then applied to $\textbf{H}$.
The adjusted hidden variables $\textbf{H}'$ are followed by the fully connected layers and the tanh layer to obtain class probability distribution $\textbf{P}=\{\textbf{p}_t\}_{t=1}^T$.

% \begin{figure}[t]
% \centering
% \includegraphics[width=\columnwidth]{figures/lstm.pdf}
% \caption{The LSTM unit in the proposed framework, where $\odot$ is the element-wise production, $\sigma$ is the sigmoid function.}
% \label{fig:lstm}
% \vspace{-3mm}
% \end{figure}

\subsection{Spatial Attention with Region Selection}

Visual attention has been shown to be an effective mechanism in image applications, by selectively focusing on certain regions in images \cite{xu2015show, ba2014multiple,mnih2014recurrent}.
In this work, we propose an attention model for region selection.
First, the inputs $\textbf{Z}^l$, as well as the $l^{\text{th}}$ deep features, are forwarded to two 2-D convolutional layers $\phi_1$ and $\phi_2$ with kernel size 1. The $\phi_1$ has both input and output channel $d$, while the $\phi_2$ has input channel $d$ and output channel $1$.
Then, the 2-D attention weights $\alpha_{t}(i, j)$ for each time step $t$ at coordinate $(i, j)$ is defined by:
\begin{equation}
\alpha_{t}(i, j) = \frac{\text{exp}(\textbf{a}_{t}(i, j))}{\sum\limits_{i}\sum\limits_{j}\text{exp}(\textbf{a}_{t}(i, j))}, {\alpha}_{t} \in {\rm I\!R}^2
\label{eq:attn_s_softmax}
\end{equation}
\begin{equation}
\begin{aligned}
\textbf{a}_{t} = \textbf{W}_{a}\tanh( & \textbf{W}^{z}_{a}{\phi_2}({\phi}_1(\textbf{z}^{l}_{t})) + \\ & \textbf{W}^{h}_{a}\textbf{h}_{t-1} + \textbf{b}^{zh}_{a}) + \textbf{b}_{a},
\end{aligned}
\label{eq:attn_s_output}
\end{equation}
where $\textbf{W}_{a}, \textbf{W}^{z}_{a}$, $\textbf{W}^{h}_{a}$, are the learnable parameter matrices, $\textbf{b}^{zh}_{a}, \textbf{b}_{a}$ are the bias vectors.
The attention weights $\alpha_t$  is a 2-D matrix where each cell spatially corresponds to the vector in $\textbf{z}^l_t$. 
A softmax selection is adopted to emphasize the corresponding regions in the latent features.
By performing element-wise product with $\alpha_t$, the weighted $\textbf{z}^{l}_t$ are forwarded to the $\text{CNN}_\text{tail}$.

\subsection{Temporal Attention with Frame Selection}

In a sequence, the input at each time $t$ contains temporal information with different importance weights for the final classification.
For example, the moment when taillights are flashing is more valuable than other frames when the network is trying to recognize the state of a vehicle's taillights.
Hence, we apply an attention model along the temporal direction to emphasize critical moments for vehicle taillight recognition.

Based on the outputs provided by the LSTM network, we integrate the attention model proposed in \cite{gehring2017convolutional} into the proposed framework.
The temporal attention $\beta_{t,u}$ corresponding to the $t^{\text{th}}$ state summary and the $u^{\text{th}}$ hidden variable is computed as a dot-product between $\textbf{d}_t$ and $\textbf{h}_{u}$, and then followed by a soft-selection:
\begin{equation}
\beta_{t,u} = \frac{\text{exp}(\textbf{d}_t \cdot \textbf{h}_{u})}{\sum_{k=1}^{T}\text{exp}(\textbf{d}_{t} \cdot \textbf{h}_{k})}.
\label{eq:attn_t_softmax}
\end{equation}
The state summary $\textbf{d}_{t}$ at time step $t$ is defined by:
\begin{equation}
\textbf{d}_{t} = \textbf{W}^{h}_{d}\textbf{h}_{t} + \textbf{W}^{c}_{d}\tanh(\textbf{c}_{t}) + \textbf{b}^{hc}_{d},
\label{eq:state_summary}
\end{equation}
where $\textbf{W}^{h}_{d}$, $\textbf{W}^{c}_{d}$ are the learnable parameter matrices, and $\textbf{b}_{d}$ is a bias vector.
This implies how the $u^{\text{th}}$ input contributes to the $t^{\text{th}}$ output.
Therefore, the output hidden variable is adjusted according to $\beta_{t,u}$:
\begin{equation}
\textbf{h}'_t = \sum_{u=1}^{T}\beta_{t,u}\textbf{h}_u.
\label{eq:attn_t_mix}
\end{equation}
The adjusted hidden variable $\textbf{h}'_t$ is then forwarded to the fully connected layers and tanh layer to obtain final prediction $\textbf{p}_{t}$ for $t^\text{th}$ time step:
\begin{equation}
\textbf{p}_t = \textbf{W}_{p}\tanh{(\textbf{W}^{h}_{p}\textbf{h}'_i + \textbf{W}^{c}_{p}\textbf{c}_{j} + \textbf{b}^{hc}_{p})} + \textbf{b}_{p},
\label{eq:attn_t_output}
\end{equation}
where $\textbf{W}_p, \textbf{W}^{h}_{p}, \textbf{W}^{c}_{p}$ are the learnable parameter matrices, and $\textbf{b}^{hc}_{p}, \textbf{b}_{p}$ are the bias vectors.

\subsection{Training Procedure}

During training, the objective loss is cross entropy loss between the predictions and labels.
To take the temporal dependence of an input sequence into account, we focus on the prediction of the last frame, i.e., $\textbf{p}_{T}$, which contains sufficient information from all the previous frames.
In other words, we compute the loss of the last frame and backpropagate all frames in the sequence with the same loss.

Due to the mutual influence of the attention models and the LSTM, the whole network needs to be optimized effectively.
First, the CNN along with LSTM network are trained from the scratch.
This allows the main stream of the network to achieve certain convergence.
Then, the CNN + LSTM network, along with the temporal attention model, are fine-tuned based on the pre-trained model in the first step.
Finally, the whole network, i.e., both spatial and temporal attention models with CNN-LSTM network, is fine-tuned from the pre-trained model in the second step. 
Such progressive training can ensure the network to converge effectively and perform better results.

%% file: 4.experiments.tex
\section{EXPERIMENTAL RESULTS}
\label{sec:experiments}

\begin{figure}[t]
    \centering
    \begin{subfigure}[b]{0.24\columnwidth}
        \centering
        \includegraphics[width=\columnwidth]{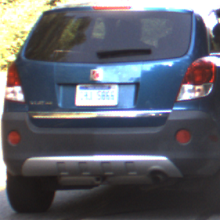}
        \vspace{-5mm}
        \caption{OOO}
    \end{subfigure}
    \begin{subfigure}[b]{0.24\columnwidth}
        \centering
        \includegraphics[width=\columnwidth]{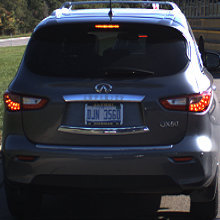}
        \vspace{-5mm}
        \caption{BOO}
    \end{subfigure}
    \begin{subfigure}[b]{0.24\columnwidth}
        \centering
        \includegraphics[width=\columnwidth]{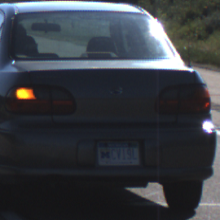}
        \vspace{-5mm}
        \caption{OLO}
    \end{subfigure}
    \begin{subfigure}[b]{0.24\columnwidth}
        \centering
        \includegraphics[width=\columnwidth]{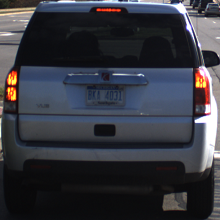}
        \vspace{-5mm}
        \caption{BLO}
    \end{subfigure}
    \\
    \begin{subfigure}[b]{0.24\columnwidth}
        \centering
        \includegraphics[width=\columnwidth]{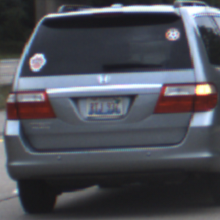}
        \vspace{-5mm}
        \caption{OOR}
    \end{subfigure}
    \begin{subfigure}[b]{0.24\columnwidth}
        \centering
        \includegraphics[width=\columnwidth]{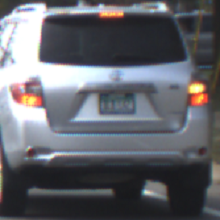}
        \vspace{-5mm}
        \caption{BOR}
    \end{subfigure}
    \begin{subfigure}[b]{0.24\columnwidth}
        \centering
        \includegraphics[width=\columnwidth]{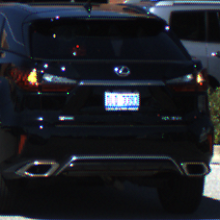}
        \vspace{-5mm}
        \caption{OLR}
    \end{subfigure}
    \begin{subfigure}[b]{0.24\columnwidth}
        \centering
        \includegraphics[width=\columnwidth]{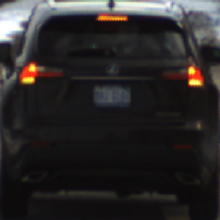}
        \vspace{-5mm}
        \caption{BLR}
    \end{subfigure}
\caption{Examples of eight taillight states in the dataset \cite{vrsd2017}.}
\label{fig:light_class}
\end{figure}

We evaluate the proposed method on UCMerced's Vehicle Rear Signal Dataset \cite{hsu2017learning}\cite{vrsd2017}, which contains 649 videos including 63,637 frames.
The sequences are recorded during the daytime under real-world driving conditions with various vehicle types.
There are eight different taillight states based on all combinations of brake and turn lights.
As shown in Figure \ref{fig:light_class}, each state is denoted by 3 letters of "B" (brake), "L" (left), and "R" (right); either the corresponding letter of the signal when it is on, or a letter O for off.
Table \ref{tab:data_distrib} shows the distribution of the states used in the experiments.

\begin{table}[t!]
\centering
\resizebox{\columnwidth}{!}{
\begin{tabular}{|c||c|c|c|c|c|c|c|c||c|}
\hline
Class & OOO & BOO & OLO & BLO & OOR & BOR & OLR & BLR & Total \\ \hline \hline
Train samples & 3256 & 2247 & 761 & 903 & 707 & 373 & 149 & 246 & 8462 \\
Test samples & 1526 & 1459 & 520 & 464 & 270 & 384 & 219 & 69 & 4811 \\
\hline
\end{tabular}
}
\caption{Distribution of the states in the train and the test sets, where a sample indicates a chunk of images.}
\label{tab:data_distrib}
\vspace{-2mm}
\end{table}

In this work, we adopt ResNet50 \cite{he2016deep} as the CNN architecture with the ImageNet pre-trained model.
The LSTM network has one recurrent layer with hidden size of 256.
The input of the network includes 128 batches each time, and each one contains 16 frames extracted from a sequence with sliding window.
A chunk of sequence which includes 16 images allows us to obtain at least one cycle of the signal state. 
Each chunk is given the ground truth label based on the video sequence
they belong to and resized to $220\times220$ pixels before feeding
to the network.
We augment the raw data with color balance, random contrast, random brightness, and horizontal flipping.
During the error backpropagation process, we compute the bootstrapped cross entropy loss \cite{reed2014training} with ratio $0.3$.
The SGD optimizer is used to adjust all parameters in our PyTorch implementation.

\begin{figure}[t]
\includegraphics[width=\columnwidth]{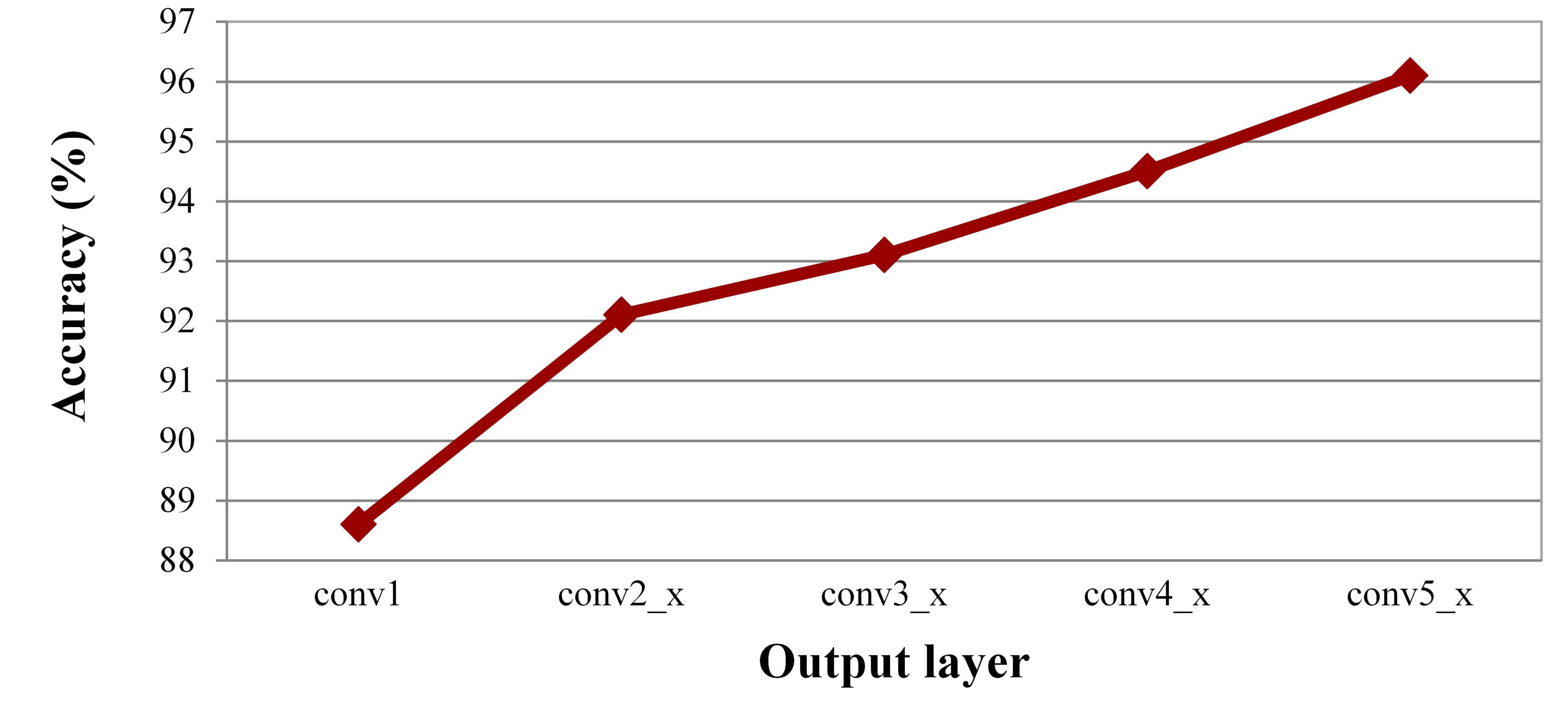}
\caption{Accuracy with different $l^\text{th}$ layers selected in the spatial attention model. Here $l$ is assigned as the integer value in each layer name. For example, the name of the $3^\text{rd}$ layer is \textit{conv3\_x}.}
\label{fig:feature_selection}
\vspace{-1mm}
\end{figure}

\begin{figure*}
    \centering
    \begin{subfigure}[t]{\textwidth}
        \centering
        \includegraphics[width=\textwidth]{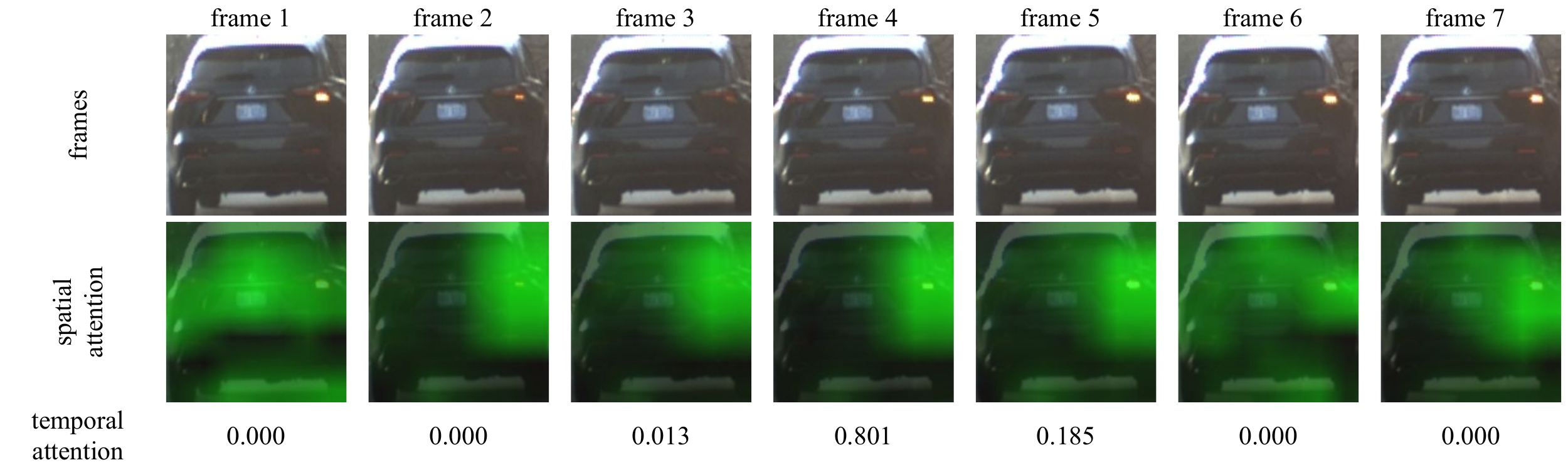}
        \vspace{-5mm}
        \caption{OOR}
        \label{fig:visualization_OOR}
    \end{subfigure}
    \hfill
    \begin{subfigure}[t]{\textwidth}
        \centering
        \includegraphics[width=\textwidth]{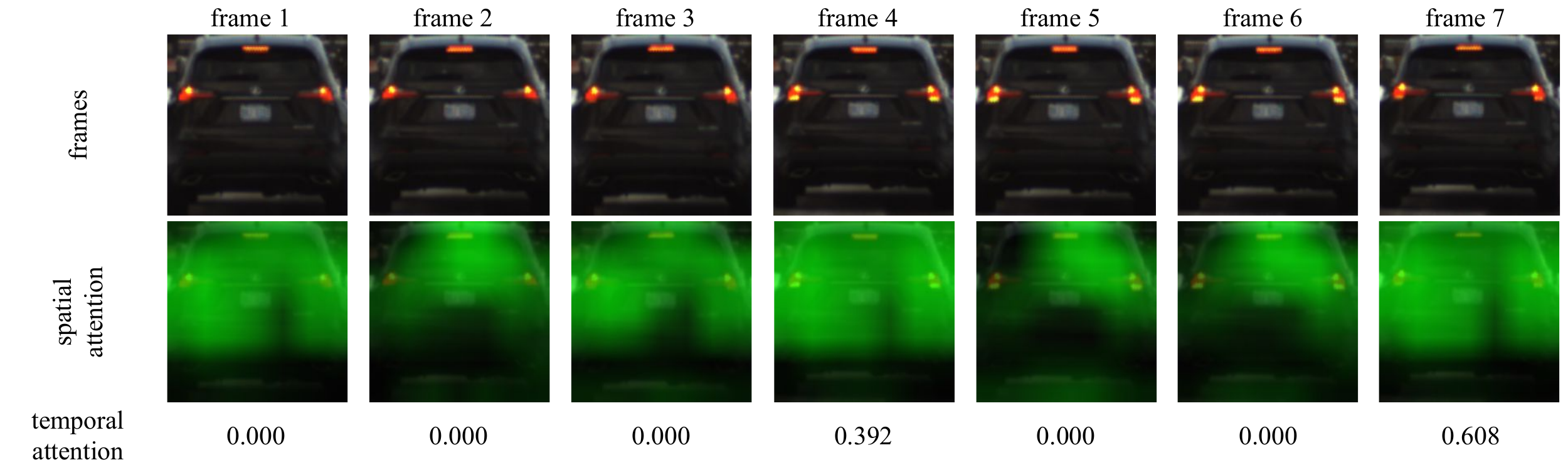}
        \vspace{-5mm}
        \caption{BLR}
        \label{fig:visualization_BLR}
    \end{subfigure}
    \vspace{-1mm}
    \caption{Visualization of the example of a) "OOR" and b) "BLR". The first rows are the input images. The second rows are the normalized spatial attentions which present intensity with green color. The temporal attention weights $\beta_{t,T}$ are listed in the bottom.}
    \label{fig:visualization}
    \vspace{-4mm}
\end{figure*}

\subsection{Quantitative Results}
\label{sec:quantitative_results}

To evaluate the performance, we compare the proposed method to the baselines, in terms of average accuracy from the predictions of all sequence chunks in each video.
The baselines we select are two 3-D CNN based methods: C3D \cite{tran2015learning} and I3D \cite{carreira2017quo}.
The I3D is embedded with Inception-v1.
Moreover, we also compare to a CNN+LSTM based method \cite{hsu2017learning}.
For fair comparison, all the inputs are the first frame and the frame difference obtained by Eq.(\ref{eq:sift_flow}), and the network settings (length of sequence chunk, hidden size, etc) remain the same.

\begin{table}[t!]
\centering
\resizebox{\columnwidth}{!}{
\begin{tabular}{|c||c|c|c|c|c|c|c|c||c|}
\hline
Method & OOO & BOO & OLO & BLO & OOR & BOR & OLR & BLR & Total \\ \hline \hline
C3D \cite{tran2015learning} & 93.1 & 89.9 & 91.7 & \textbf{95.3} & 96.3 & 96.4 & \textbf{99.5} & \textbf{100} & 93.2 \\
I3D \cite{carreira2017quo} & 96.3 & 93.1 & 94.2 & 85.1 & \textbf{97.8} & 95.1 & 98.2 & \textbf{100} & 94.2 \\
CNN-LSTM \cite{hsu2017learning} & 97.5 & 90.4 & \textbf{95.8} & 88.4 & 95.2 & 87.0 & 89.5 & 92.8 & 93.0 \\
\hline
Proposed (S) & 96.8 & 92.9 & 95 & 89.2 & 95.9 & 87.8 & 94.5 & \textbf{100} & 93.9\\
Proposed (T) & 98.4 & 95.1 & 90.8 & 89.2 & 97.0 & 92.4 & 90.0 & \textbf{100}& 94.8 \\
Proposed (S+T) & \textbf{98.6} & \textbf{95.3} & 90.8 & 93.3 & 97.0 & \textbf{97.0} & 97.7
& 98.6 & \textbf{96.1} \\
\hline
\end{tabular}
}
\caption{Quantitative results of the comparisons between the baselines. The proposed method (S)/(T) means that only spatial/temporal attention is equipped, and (S+T) means that both attention models are equipped.}
\label{tab:performance}
\vspace{-4mm}
\end{table}

Table \ref{tab:performance} shows the quantitative results of the comparisons.
The proposed (S) method is better ($+0.9\%$) than the CNN-LSTM.
This means that spatial attention model improves the performance.
The proposed (T) method performs much better than the the CNN-LSTM and 3-D CNN based methods.
This implies that the temporal attention model can encode the temporal dependencies better than the 3-D convolution.
When equipped with both spatial and temporal attention models, the proposed (S+T) method outperforms the baseline methods with $96\%$ accuracy.
Overall, facilitated by the attention models, the proposed method is able to tell the taillight states effectively.

\subsection{Feature Selection in Spatial Attention}

In spatial attention model, extracting features $\textbf{z}^{l}_t$ from different layers influence the performance because of different spatial dimensions.
To evaluate the influence, we remain the same settings used in Section \ref{sec:quantitative_results}, but input different $l^\text{th}$ layers' outputs to the spatial attention model.
Figure \ref{fig:feature_selection} shows the accuracy performance with different outputs of the $l^\text{th}$ layers, with corresponding layer names in the ResNet50.
The results show that the performance becomes better when selecting deeper features.
The performance becomes worse than without spatial attention model when selecting the \textit{conv\_1x} and the \textit{conv\_2x} layers.
This implies that paying attention to more abstract (deeper) features is more effective than paying attention to less abstract features.

\subsection{Visualization}

To analyze how the spatio-temporal attention model works in the framework, we visualize the attention weights during inference. 
For spatial attention, we normalize the weights in 2-D and visualize intensity with green color.
For spatial attention model, we select $l = 5$, i.e., the outputs of the \textit{conv5\_x} layer, as deep features.
This means that $\alpha_t$ is a $7 \times 7$ matrix.
To properly visualize the attention weights, we resize $\alpha_t$ and blend with input frames.

Figure \ref{fig:visualization_OOR} shows another example of the "OOR" case, where only the right-turn signal is flashing.
From frame 1, the spatial attention is uniformly distributed on the rear of the vehicle.
When the right-turn signal gradually switches off from frame 2 to frame 3, spatial attention starts to concentrate on the region of the right-turn signal until the last frame.
Then, the temporal attention pays much attention to the frame 4, while the right-turn signal is on at this time step.
Meanwhile, the spatial attention weight goes up to $0.8$, which implies that the network pays more attention to significant changes of the signal.

Figure \ref{fig:visualization_BLR} shows another example of the "BLR" case, where the brake signal is on and both turn signals are flashing.
As shown in the figure, the spatial attention focuses on both sides of the vehicle.
Both left- and right-turn signals start to turn on at frame 4.
This triggers the network to pay temporal attention to the frame 4.
The signals become off until the frame 7, while the temporal attention weight goes up to $0.6$, which is higher than that in frame 4.
This may be caused by the network learning a cycle of signal flashing for this chunk of sequence.

\begin{figure}[t]
    \centering
    \begin{subfigure}[t]{0.24\columnwidth}
        \centering
        \includegraphics[width=\columnwidth]{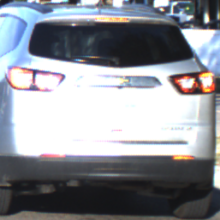}
        \vspace{-5mm}
        \caption{BOO / BLR}
        \label{fig:BOO_BLR}
    \end{subfigure}
    \begin{subfigure}[t]{0.48\columnwidth}
        \centering
        \includegraphics[width=\columnwidth]{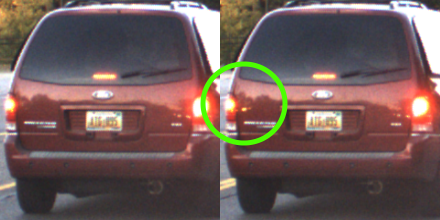}
        \vspace{-5mm}
        \caption{BOO / BLO}
        \label{fig:BOO_BLO}
    \end{subfigure}
    \begin{subfigure}[t]{0.24\columnwidth}
        \centering
        \includegraphics[width=\columnwidth]{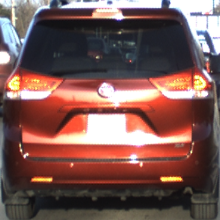}
        \vspace{-5mm}
        \caption{BOO / OOO}
        \label{fig:BOO_OOO}
    \end{subfigure}
    \\
    \begin{subfigure}[b]{0.24\columnwidth}
        \centering
        \includegraphics[width=\columnwidth]{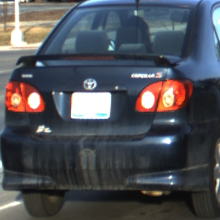}
        \vspace{-5mm}
        \caption{OOO / BOO}
        \label{fig:OOO_BOO}
    \end{subfigure}
    \begin{subfigure}[b]{0.48\columnwidth}
        \centering
        \includegraphics[width=\columnwidth]{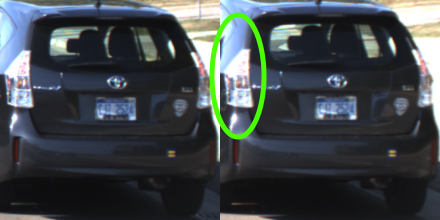}
        \vspace{-5mm}
        \caption{OOO / OLO}
        \label{fig:OOO_OLO}
    \end{subfigure}
    \begin{subfigure}[b]{0.24\columnwidth}
        \centering
        \includegraphics[width=\columnwidth]{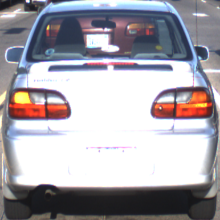}
        \vspace{-5mm}
        \caption{OLO / BLO}
        \label{fig:OLO_BLO}
    \end{subfigure}
\caption{Failure cases in the test set, where each subcaption indicates prediction/label.}
\label{fig:failure_cases}
\vspace{-4mm}
\end{figure}

\subsection{Failure Cases}

Figure \ref{fig:failure_cases} shows several failure cases in the test set.
Most failure cases are due to lighting reflection which makes signal-off recognized as the signal-on, as the examples show in Figure \ref{fig:BOO_BLR}, \ref{fig:BOO_OOO}, \ref{fig:OOO_BOO}, and \ref{fig:OLO_BLO}.
A few edge cases shown in Figure \ref{fig:BOO_BLO} and \ref{fig:OOO_OLO} are false alarms, where the signal-off is treated as flashing signals.
These are due to the reflection of the other vehicle's flashing light, or a sudden change of sunlight reflection.

%% file: 5.conclusions.tex
\section{CONCLUSIONS}
\label{sec:conclusions}

We propose a method that learns to recognize eight different taillight states from a video sequence.
Our method integrates both spatial and temporal attention models into an LSTM network, so as to effectively exploit deep features in both spatial and temporal domains.
The experimental results show that the proposed method is able to effectively predict each taillight states in real-world traffic scenes.